\documentclass[letterpaper, 10 pt, conference]{ieeeconf}  
\IEEEoverridecommandlockouts                              
\usepackage{stfloats}
\usepackage{bm,bbm,float,booktabs}
\usepackage{multirow}
\usepackage{graphicx}
\usepackage{amssymb,amsfonts,amsmath}

\usepackage{amsthm}
\usepackage{algorithm,algorithmic}
\newtheorem{theorem}{Theorem}

\newtheorem{definition}{Definition}

\title{\LARGE \bf Exploring Beyond-Demonstrator via Meta Learning-Based Reward Extrapolation}
\author{Mingqi Yuan$^{1}$ and Man-on Pun$^{2}$
\thanks{*This work was supported by National Key Research and Development Program of China under Grant No. 2020YFB1807700. (Corresponding author: Man-On Pun.)}
\thanks{$^{1}$Mingqi Yuan is with the School of Science and Engineering, The Chinese University of Hong Kong, Shenzhen, 518172 China, and also with Shenzhen Research Institute of Big Data, Shenzhen, 518172 China
        {\tt\small mingqiyuan@link.cuhk.edu.cn}}%
\thanks{$^{2}$Man-on Pun is with the School of Science and Engineering, The Chinese University of Hong Kong, Shenzhen, 518172 China, and also with Shenzhen Research Institute of Big Data, Shenzhen, 518172 China
        {\tt\small simonpun@cuhk.edu.cn}}%
}

\begin{document}
\maketitle
\thispagestyle{empty}
\pagestyle{empty}
\begin{abstract}
Extrapolating beyond-demonstrator (BD) performance through the imitation learning (IL) algorithm aims to learn from and subsequently outperform the demonstrator. To that end, a representative approach is to leverage inverse reinforcement learning (IRL) to infer a reward function from demonstrations before performing RL on the learned reward function. However, most existing reward extrapolation methods require massive demonstrations, making it difficult to be applied in tasks of limited training data. To address this problem, one simple solution is to perform data augmentation to artificially generate more training data, which may incur severe inductive bias and policy performance loss. In this paper, we propose a novel meta learning-based reward extrapolation (MLRE) algorithm, which can effectively approximate the ground-truth rewards using limited demonstrations. More specifically, MLRE first learns an initial reward function from a set of tasks that have abundant training data. Then the learned reward function will be fine-tuned using data of the target task. Extensive simulation results demonstrated that the proposed MLRE can achieve impressive performance improvement as compared to other similar BDIL algorithms. Our code is available at GitHub\footnote{https://github.com/yuanmingqi/MLRE}.
\end{abstract}

\section{INTRODUCTION}
Imitation learning (IL) aims to recover an expert policy from demonstrations of a specific task. IL is very effective for solving complex tasks with minimal expert knowledge when it is simpler for an expert to demonstrate the expected behavior \cite{hussein2017imitation}. The simplest form of IL is behavioral cloning that straightforwardly learns the mapping relationship from observations to actions using supervised learning. However, behavioral cloning requires massive demonstration data and suffers from the compounding error problem, {\em i.e.}, the learned policy may be invalid if the data distribution is vastly distinct from the training set. Alternatively, inverse reinforcement learning (IRL) was leveraged to first learns a reward function from demonstrations before performing RL with the inferred reward function \cite{ng2000algorithms}. However, the recovered policy has been consistently found sub-optimal, failing to outperform the demonstrator as IRL is designed to find the reward function making the demonstrations appear optimal.

Learning from and outperforming the demonstrator via IL is commonly referred to as beyond-demonstrator (BD) IL in the literature. The concept of extrapolating BD performance from demonstrations was first proposed in \cite{brown2019extrapolating} by designing a trajectory-ranking-reward-extrapolation (TREX) framework. TREX first collects a series of ranked trajectories before training a parameterized reward function that matches the rank relation. After that, the reward function is employed to learn a policy via RL. By fully excavating the rank information, TREX can accurately approximates the ground-truth reward function to learn BD polices. In particular, TREX was further extended to the multi-agent task in \cite{huang2020ma}. However, it is always difficult to get well-ranked trajectories in real-world scenarios. To address this problem, \cite{brown2020better} proposed a disturbance-based-reward-extrapolation (DREX) framework to automatically generate the ranked demonstrations. However, it was found that DREX incorrectly assumes an ordinal and homogeneous noise-performance relationship across the noise-injected policies, resulting in severe learning errors \cite{chen2020learning}.

To reduce the dependency of demonstrations, \cite{yu2020intrinsic} proposed an intrinsic-reward-driven-imitation-learning (GIRIL) framework, which only takes a one-life demonstration to learn a family of reward functions using variational autoencoder (VAE) \cite{kingma2013auto}. In particular, \cite{yu2020intrinsic} first introduced the intrinsic reward to IL to explore the BD policies. In sharp contrast to the rewards explicitly given by the environment, intrinsic rewards characterize the inherent learning motivation of the agent. Extensive experiments demonstrated that intrinsic rewards could significantly improve the exploration of the environment and result in higher performance, even in complex environments with high-dimensional observations \cite{burda2018large}. However, despite its many advantages, GIRIL suffers from poor interpretability and low robustness as the intrinsic rewards may have less correlation with the ground-truth rewards. Moreover, the excessive exploration may lead to the television dilemma reported in \cite{savinov2018episodic}. Finally, the one-life demonstration configuration is delicate that heavily depends on the quality of the collected demonstration.

Inspired by the discussions above, we consider developing a few-shot reward extrapolation framework to learn high-quality reward functions based on limited demonstrations. Our key insight is to fully extract and exploit the original information of the demonstrations via meta learning, which aims to \textit{learn to learn} and effectively solves the few-shot learning problem. Our main contributions are summarized as follows:
\begin{itemize}
\item We propose a meta learning-based reward extrapolation (MLRE) algorithm that overcomes the problem of limited demonstrations. MLRE first learns an initial reward function from a set of training tasks that have abundant training data. Then, the learned reward function will be fine-tuned using data of the target task. In addition, we improve the loss function of the trajectory-ranking method. We demonstrate that MLRE can accurately approximate the ground-truth rewards even with fewer demonstrations.

\item Extensive simulation is performed to compare the policy performance of MLRE against existing methods using Atari games with high-dimensional observations. Simulation results confirm that the proposed method achieves superior performance with higher efficiency and robustness.
\end{itemize}

\section{PROBLEM FORMULATION}
In this paper, we study the BDIL problem considering the Markov decision process (MDP) defined by a tuple $\mathcal{M}=\langle \mathcal{S},\mathcal{A},P,R^{*},\rho({\bm s}_0),\gamma\rangle$, in which $\mathcal{S}$ is the state space, $\mathcal{A}$ is the action space, $P({\bm s}'|{\bm s},{\bm a})$ is the transition probability, $R^{*}:\mathcal{S}\times\mathcal{A}\rightarrow\mathbb{R}$ is the ground-truth reward function, $\rho({\bm s}_0)$ is the initial state distribution, and $\gamma\in(0,1]$ is a discount factor. Note that $R^{*}$ is solely determined by the task, and the performance of agent is only evaluated by $R^{*}$. Finally, we denote by $\pi({\bm a}|{\bm s})$ the policy of the agent that selects an action from the action space based on the state of the environment. Equipped with these definitions, we first define the objective of RL:
\begin{equation}\label{eq:rl objective}
\pi^{*}=\underset{\pi\in\Pi}{\rm argmax}\;J(\pi|R^{*}),
\end{equation}
where $J(\pi|R^{*})=\mathbb{E}_{\tau\sim\pi}\sum_{t=0}^{T-1}\gamma^{t}R^{*}_{t}({\bm s}_t,{\bm a}_t)$, $\Pi$ is the set of all possible stationary policies, and $\tau=({\bm s}_{0},{\bm a}_{0},\dots,{\bm a}_{T-1},{\bm s}_{T})$ is the trajectory collected by the agent.

In contrast, IL aims to learn a {\em generation} policy $\hat{\pi}$ that can provide comparable performance as a given demonstrator. Denote by $\mathcal{D}=\{\tau_{1},\dots,\tau_{N}\}$ the set of demonstrations, the objective of IL can be defined as a reduction to maximum likelihood estimation (MLE):
\begin{equation}
\hat{\pi}=\underset{\pi\in\Pi}{\rm argmax}\sum_{(\bm{s},\bm{a})\in\tau,\tau\in\mathcal{D}}\log \pi(\bm{a}|\bm{s}).
\end{equation}

In this paper, we aim to learn a BD policy through IL, which requires the agent to imitate and outperform the demonstrator. Mathematically, such an objective can be defined as follows:
\begin{definition}
	Given a set of demonstrations $\mathcal{D}=\{\tau_{1},\dots,\tau_{N}\}$ drawn from a demonstrator, BDIL aims to learn a generation policy $\hat{\pi}$ based on $\mathcal{D}$, such that
	\begin{equation}
	J(\hat{\pi}|R^{*})>J(\mathcal{D}|R^{*})=\frac{1}{|\mathcal{D}|}\sum_{\tau\in\mathcal{D}}J(\tau|R^{*}),
	\end{equation}
	where $J(\tau|R^{*})=\sum_{({\bm s}_t,{\bm a}_t)\in\tau}\gamma^{t}R^{*}({\bm s}_t,{\bm a}_t)$, $J(\mathcal{D}|R^{*})$ is the estimation of the expected discounted return of the demonstrator policy.
\end{definition}

Clearly, it is analytically intractable to derive the optimal generation policy via simple imitations. In the following sections, we first demonstrate a theoretical justification of the BD objective before proposing a novel and efficient algorithm to learn a BD policy.

\section{THEORETICAL JUSTIFICATION OF BDIL}
Considering an IRL scenario, whose objective is to learn the reward function of the demonstrator and then use it to optimize a policy. A common approach is to represent the reward function as a linear combination of features:
\begin{equation}
R(\bm{s})=\bm{w}^{T}\phi(\bm{s}),
\end{equation}
where $\bm{w}$ is a weighting vector and $\phi(\cdot)$ is an encoding function.

The expected return of a policy evaluated by $R(\bm{s})$ is given by:
\begin{equation}
J(\pi|R)=\bm{w}^{T}\mathbb{E}_{\pi}\left[ \sum_{t=0}^{\infty}\gamma^{t}\phi(\bm{s}_{t}) \right]=\bm{w}^{T}\bm{\Phi}_{\pi}.
\end{equation}

The following theorem provides a theoretical condition for realizing the BD objective:
\begin{theorem}\label{thm:bd}
	If the estimated reward function is $\hat{R}(\bm{s})=\bm{w}^{T}\phi(\bm{s})$, the true reward function is $R^{*}(\bm{s})=\hat{R}(\bm{s})+\epsilon(\bm{s})$ for an error function $\epsilon:\mathcal{S}\rightarrow\mathbb{R}$ and $\Vert\bm{w}\Vert_{1}\leq 1$, then extrapolating BD policy is guaranteed if:
	\begin{equation}
	J(\pi^{*}|R^{*})-J(\mathcal{D}|R^{*})>\epsilon_{\bm{\Phi}}+\frac{2\Vert\epsilon\Vert_{\infty}}{1-\gamma}
	\end{equation}
	where $\epsilon_{\bm{\Phi}}=\Vert\bm{\Phi}_{\pi^{*}}-\bm{\Phi}_{\hat{\pi}}\Vert_{\infty}$, $\pi^{*}$ is the optimal policy under $R^{*}$, $\hat{\pi}$ is the generation policy, and $\Vert\epsilon\Vert_{\infty}=\sup\{ |\epsilon(\bm{s})|:\bm{s}\in\mathcal{S} \}$.
\end{theorem}
\begin{proof}
	See proof in \cite{brown2020better}.
\end{proof}
To extrapolate a BD policy, Theorem~\ref{thm:bd} indicates that the demonstrator should be sufficiently suboptimal, and the error of the learned reward function should be sufficiently small. In particular, the generation policy has to approximate the optimal policy as accurate as possible. Therefore, our objective is to precisely recover the ground-truth reward function, and the RL can guarantee that $\epsilon_{\bm{\Phi}}$ is small.

\section{META LEARNING-BASED REWARD EXTRAPOLATION}
In this paper, we learn the reward function following the trajectory-ranking approach proposed in \cite{brown2019extrapolating}. Given a sequence of $N$ ranked demonstrations $\tau_{1}\prec\tau_{2}\prec,\dots,\prec\tau_{N}$, TREX performs reward inference using a neural network $\hat{R}_{\bm\theta}(\bm{s})$, such that
\begin{equation}
\sum_{\bm{s}\in\tau_{i}}\hat{R}_{\bm\theta}(\bm{s})<\sum_{\bm{s}\in\tau_{j}}\hat{R}_{\bm\theta}(\bm{s}),
\end{equation}
where $\tau_{i}\prec\tau_{j}$. The reward function is learned by minimizing a pairwise ranking loss as follows:
\begin{equation}
L_{\rm RE}(\bm{\theta},\mathcal{P})=-\frac{1}{|\mathcal{P}|}\sum_{(i,j)\in\mathcal{P}}\log \frac{S(j)}{S(i)+S(j)},
\end{equation}
where $\mathcal{P}=\{(i,j):\tau_{i}\prec\tau_{j}\}$, $S(i)=\exp\{\sum_{\bm{s}\in\tau_{i}}\hat{R}_{\bm\theta}(\bm{s})\}$. After that, the derived reward function can be combined with any RL algorithms to learn a policy.

However, learning an accurate reward function via trajectory-ranking requires massive high-quality demonstrations, which is impractical in real-world scenarios. Furthermore, self-generated demonstrations amy introduce detrimental inductive bias. To address this problem, we introduce the following meta learning method to realize efficient reward extrapolation with limited demonstrations. Traditional supervised learning methods let the model recognize the samples in the training set and then generalize to the test set. In contrast, meta learning aims to \textit{learn to learn} and effectively solve the few-shot learning problem \cite{finn2017model}.

\subsection{Meta Learning}
Consider a model $f_{\bm\theta}$ represented by a neural network with parameters ${\bm\theta}$, which maps observations $\bm{x}$ to outputs $\bm{y}$. Meta learning aims to train this model to be able to adapt to a set of tasks. Each task can be defined as a tuple $\mathcal{T}=\langle L(\bm{x}_{0},\bm{y}_{0},\dots,\bm{x}_{T},\bm{y}_{T}),q(\bm{x}_{0}),q(\bm{x}_{t+1}|\bm{x}_{t},\bm{y}_{t}), T  \rangle$, where $L$ is a loss function, $q(\bm{x}_{0})$ is an initial distribution, $q(\bm{x}_{t+1}|\bm{x}_{t},\bm{y}_{t})$ is a transition distribution, and $T$ is an episode length. In particular, the episode length is one for independent identically distributed supervised learning. Furthermore, we denote by $p(\mathcal{T})$ the distribution of tasks that we want the model to adapt to. During meta-training, we first sample a new task $\mathcal{T}_{i}$ from $p(\mathcal{T})$ before training the model with training data $\mathcal{T}_{i}$. After that, the model is improved by evaluating the test error with respect to the parameters, which serves as the training error of the meta-learning process. These procedures are repeated for multiple times before the learned parameters are saved. Finally, we can perform fine-tuning on the learned parameters to adapt to our target task.

\subsection{MLRE}
In this section, we propose a meta learning-based reward extrapolation (MLRE) framework. Our key insight is to fully exploit the original information extracted from the demonstrations to recover high-quality reward functions via meta learning. Our reward extrapolation task can be defined as
\begin{equation}
\mathcal{T}=\langle \mathcal{D}, L_{\rm RE}\rangle.
\end{equation}
Moreover, we redefine the pairwise ranking loss as follows:
\begin{equation}
	\begin{aligned}
		L_{\rm RE}(\bm{\theta},\mathcal{P})&=-\frac{1}{|\mathcal{P}|}\sum_{(i,j)\in\mathcal{P}}\bigg[\log \frac{S(j)}{S(i)+S(j)}\\
		&+\left|\frac{\mathrm{Len}(\tau_{i})}{S(i)}-\lambda\right|+\left|\frac{\mathrm{Len}(\tau_{j})}{S(j)}-\lambda\right|\bigg],
	\end{aligned}
\end{equation}
where $\mathrm{Len}(\tau_{i})$ is the length of $\tau_{i}$ and $\lambda>0$ is a scaling coefficient. The regularization term indicates that the agent can get higher scores if it lives longer. Moreover, it can limit the output range of the learned reward function.

\begin{figure}[h]
	\centering
	\includegraphics[width=\linewidth]{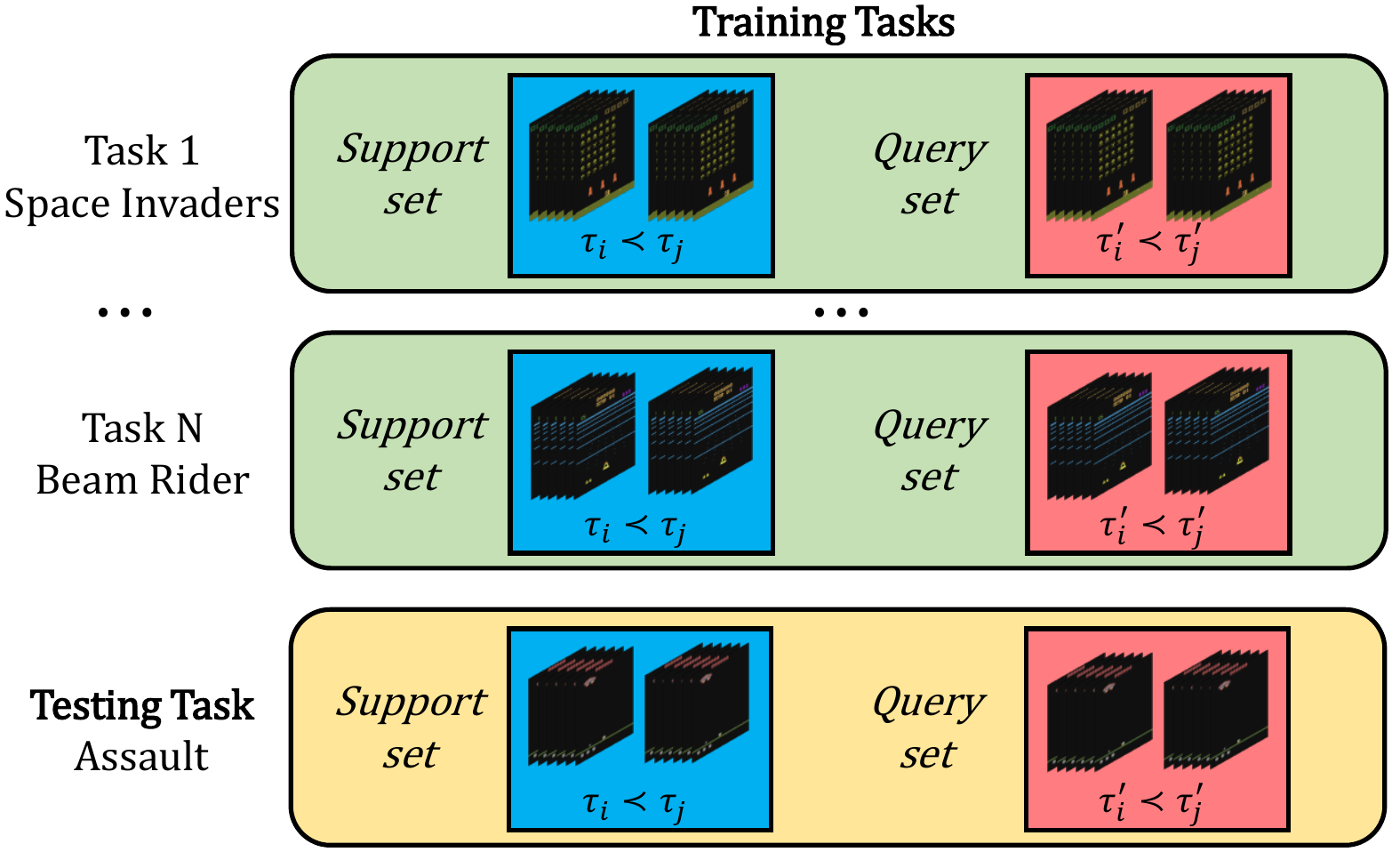}
	\caption{An example of the training tasks and the testing task.}
	\label{fig:task}
\end{figure}

To perform meta learning, several training tasks are required with each task containing a support set (training data) and a query set (testing data). Take the Atari games for instance, we want to recover the reward function of the \textit{Assault} game shown in Fig.~\ref{fig:task}. However, we only have few demonstrations drawn from a trained demonstrator. Fortunately, there are some demonstrations of other games, such as \textit{Beam Rider} and \textit{Space Invaders}, which have similar playing methods and reward mechanisms. Therefore, \textit{Beam Rider} and \textit{Space Invaders} are set as the training tasks, and \textit{Assault} is set as the testing task. Equipped with additional demonstrations from the other two games, we can leverage meta learning to learn a better reward function for the \textit{Assault} game.

\begin{figure*}[ht]
	\centering
	\includegraphics[width=0.9\linewidth]{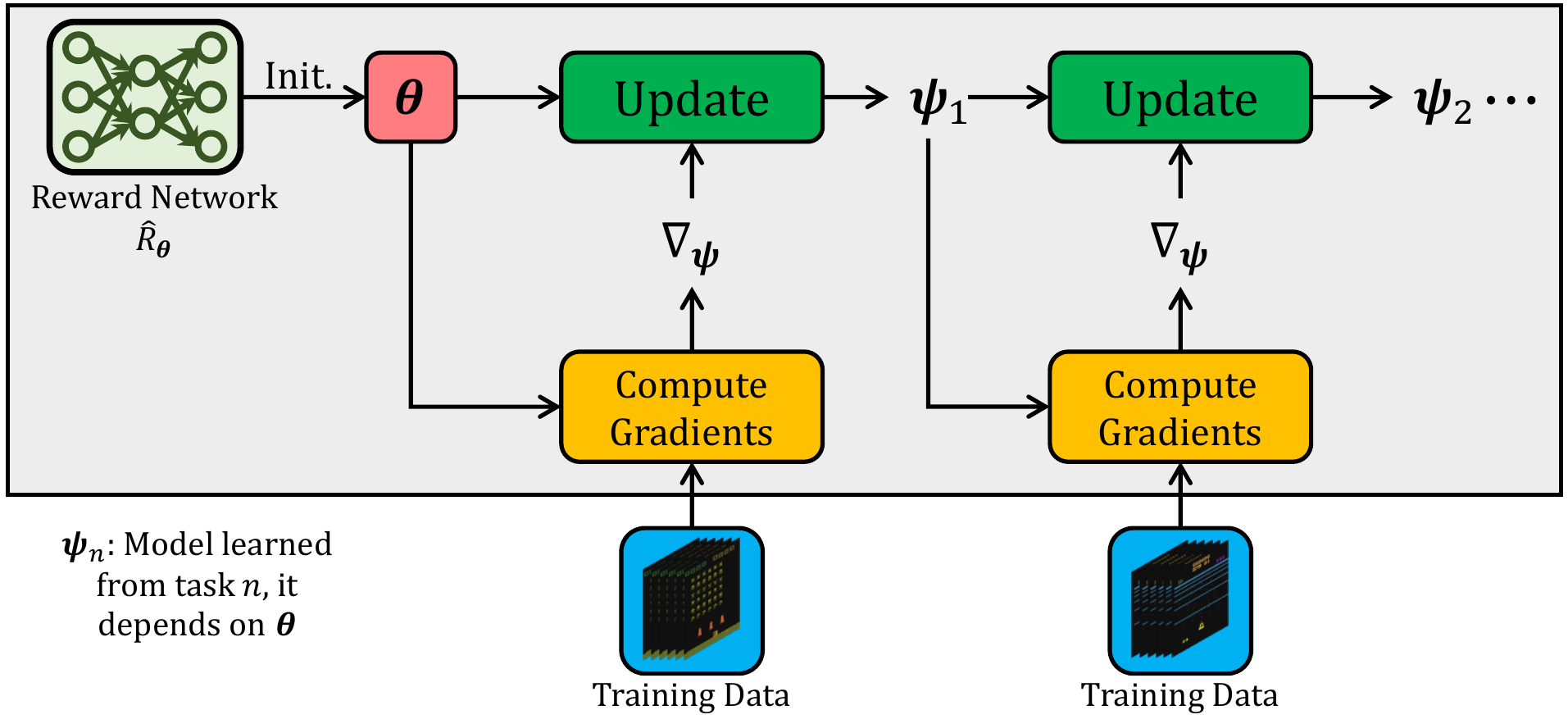}
	\caption{The overview of the MLRE framework.}
	\label{fig:mlre}
\end{figure*}

MLRE is designed using a model-agnostic meta learning (MAML) method in \cite{finn2017model}. Recall the parameterized reward function $\hat{R}_{\bm\theta}$, and sample a new task $\mathcal{T}_{i}$ from $p(\mathcal{T})$. When the reward network $\hat{R}_{\bm\theta}$ adapts to the new task, its parameters change from $\bm{\theta}$ to $\bm{\theta}'_{i}$. MAML computes $\bm{\theta}'_{i}$ using one or multiple gradients with respect to task $\mathcal{T}_{i}$. For one-step update, we have
\begin{equation}\label{eq:gradient descent}
\bm{\theta}'_{i}=\bm{\theta}-\alpha \nabla_{\bm\theta} L_{\rm RE}(\bm{\theta},\mathcal{P}_{i}),
\end{equation}
where $L_{\rm RE}$ is evaluated on the demonstrations of task $\mathcal{T}_{i}$, and $\alpha$ is a step size. Finally, the model parameters are trained by minimizing the following loss function across from tasks sampled from $p(\mathcal{T})$:
\begin{equation}
L_{\rm Meta}=\sum_{\mathcal{T}_{i}\sim p(\mathcal{T})}L_{\rm RE}(\bm{\theta}_{i}',\mathcal{P}_{i}).
\end{equation}
Using the stochastic gradient descent, the model parameters are updated as follows:
\begin{equation}\label{eq:update meta}
\bm{\theta}\leftarrow\bm{\theta}-\beta\nabla_{\bm\theta}L_{\rm Meta},
\end{equation}
where $\beta$ is the meta step size. Equipped with the learned reward function, any RL algorithms can be used to learning a policy. We illustrate the complete workflow of MLRE in Fig.~\ref{fig:mlre}. In practice, we maintain an independent model $\bm{\psi}_{n}$ for the $n$-th task that has identical architecture with $\hat{R}_{\bm\theta}$, and let $\bm{\psi}_{0}=\bm{\theta}$. During the meta-training, we only focus on the initialization parameters $\bm{\theta}$. Finally, we summarize the full algorithm of MLRE in Algorithm \ref{algo:mlre}.

\begin{algorithm}[htp]
	\caption{MLRE}\label{algo:mlre}
	\begin{algorithmic}[1]
	\STATE Collect demonstrations $\mathcal{D}$;
	\STATE Randomly initialize the reward network $\hat{R}_{\bm\theta}$;
	\STATE Initialize a set of training tasks;
	\STATE Initialize the step size hyper-parameters $\alpha,\beta$;
	\WHILE {not done}
	\STATE Sample batch of tasks $\mathcal{T}_{i}\sim p(\mathcal{T})$;
	\FOR {all $\mathcal{T}_{i}$}
	\STATE Construct training dataset $\mathcal{P}_{i}$ using $\mathcal{D}_{i}$;
	\STATE Evaluate $\nabla_{\bm\theta} L_{\rm RE}(\bm{\theta},\mathcal{P}_{i})$ with respect to  $\mathcal{P}_{i}$;
	\STATE Compute adapted parameters with gradient descent using Eq.~\eqref{eq:gradient descent};
	\ENDFOR
	\STATE Update the reward network using Eq.~\eqref{eq:update meta};
	\ENDWHILE
	\STATE Optimize the generation policy $\hat{\pi}$ via any RL algorithms on the learned reward function.
	\end{algorithmic}
\end{algorithm}

\section{EXPERIMENTS}
In this section, we evaluate the MLRE on six Atari games of OpenAI Gym library, namely \textit{Assault}, \textit{Battle Zone}, \textit{Kung Fu Master}, \textit{Phoenix}, \textit{Riverraid}, and \textit{Space Invaders}. For benchmarking, several most representative algorithms are carefully selected, namely GIRIL, DREX, and Wasserstein adversarial imitation learning (WAIL) \cite{xiao2019wasserstein}. The first two methods are BDIL algorithms, while the latter is an IL algorithm. With WAIL, we can validate that the MLRE can imitate and outperform the demonstrator. With GIRIL and DREX, we can validate that the MLRE can realize higher performance with higher efficiency and robustness. As for hyper-parameters setting, we only report the values of the best experiment results.

\subsection{Demonstrations}
To generate suboptimal demonstrations, we trained a proximal policy optimization (PPO) agent using the ground-truth reward for ten million steps \cite{schulman2017proximal}. More specifically, we used a PyTorch implementation of the PPO created by \cite{kostrikov2018github} with its default hyper-parameters. After that, we generate $50$ one-life demonstrations using the trained PPO agent for all the games. A one-life demonstration only has the states and actions performed by the demonstrator until it dies for the first time in a game, while the full-episode demonstration is derived after demonstrator losing all available lives. Therefore, the one-life demonstration data is more limited and challenging for reward extrapolation.

\begin{table}[h]
	\caption{The architecture of the modules.}
	\label{tb:cnn na}
	\centering
	\begin{tabular}{l|ll}
		\hline
		\textbf{Module} & \multicolumn{1}{l|}{Policy network}                                                                                                                                                            & Value network                                                                                                                                            \\ \hline
		Input  & \multicolumn{1}{l|}{States}                                                                                                                                                                    & States                                                                                                                                                   \\ \hline
		Arch.  & \multicolumn{1}{l|}{\begin{tabular}[c]{@{}l@{}}8$\times$8 Conv 32, ReLU\\ 4$\times$4 Conv 64, ReLU\\ 3$\times$3 Conv 32, ReLU\\ Flatten \\Dense 512\\ Categorical Distribution\end{tabular}} & \begin{tabular}[c]{@{}l@{}}8$\times$8 Conv 32, ReLU\\ 4$\times$4 Conv 64, ReLU\\ 3$\times$3 Conv 32, ReLU\\ Flatten \\Dense 512\\ Dense 1\end{tabular} \\ \hline
		Output & \multicolumn{1}{l|}{Actions}                                                                                                                                                                   & Predicted values                                                                                                                                         \\ \hline
		\textbf{Module} & \multicolumn{2}{c}{Reward function}                                                                                                                                                                                                                                                                                                                       \\ \hline
		Input  & \multicolumn{2}{c}{States}                                                                                                                                                                                                                                                                                                                                \\ \hline
		Arch.  & \multicolumn{2}{c}{\begin{tabular}[c]{@{}c@{}}8$\times$8 Conv 32, ReLU\\ 4$\times$4 Conv 64, ReLU\\ 3$\times$3 Conv 32, ReLU\\ Flatten\\ Dense 512, ReLU\\ Dense 1\end{tabular}}                                                                                                                                                                          \\ \hline
		Output & \multicolumn{2}{c}{Estimated rewards}                                                                                                                                                                                                                                                                                                                     \\ \hline
	\end{tabular}
\end{table}

\begin{figure*}[ht]
	\includegraphics[width=\linewidth]{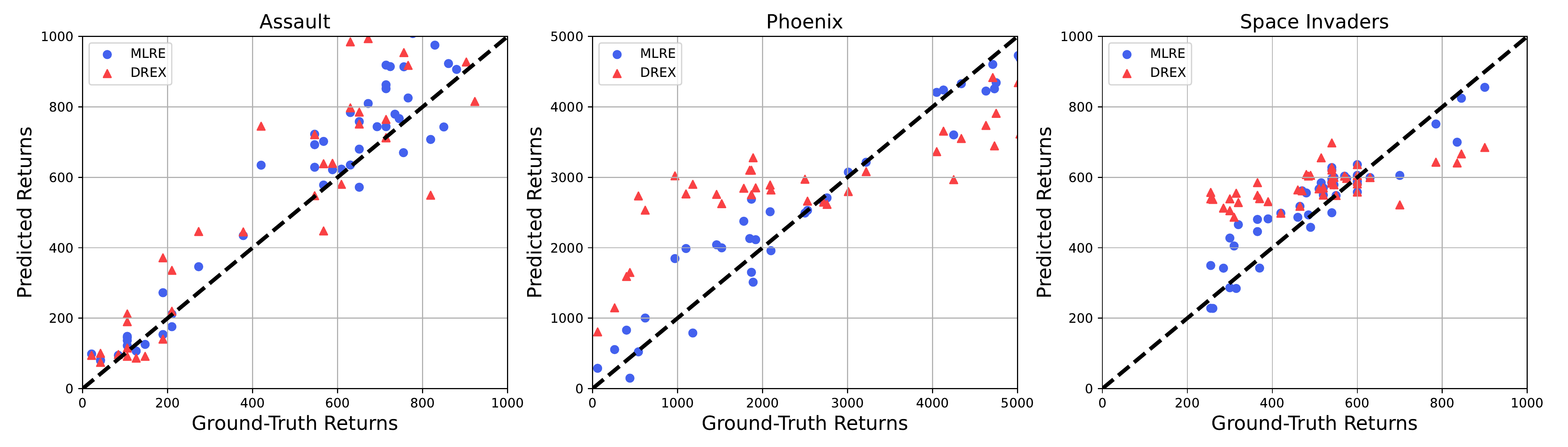}
	\caption{Reward extrapolation for three Atari games. The black dashed line represents the performance range of the demonstrator.}
	\label{fig:re}
\end{figure*}

\begin{figure*}[hb]
	\centering
	\includegraphics[width=0.95\linewidth]{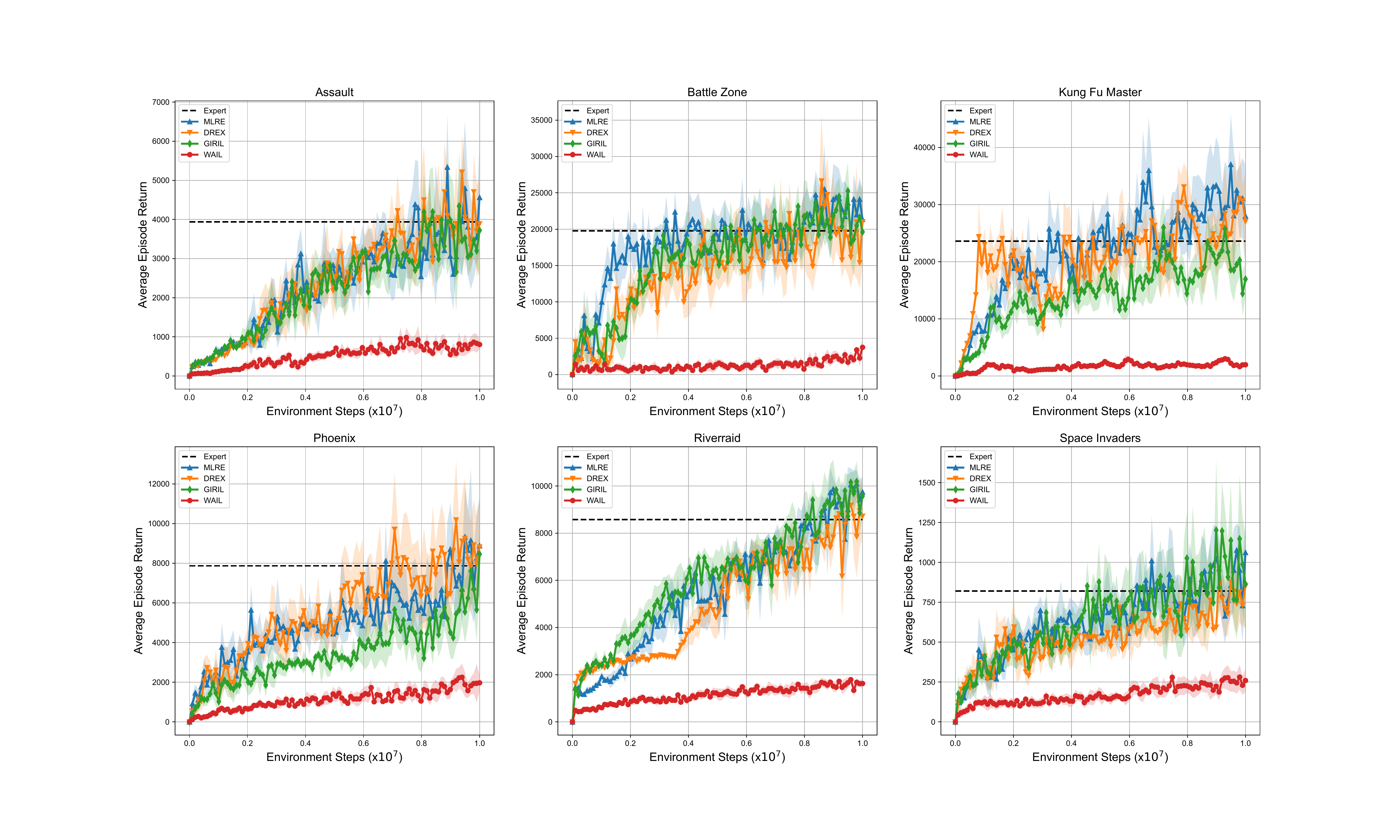}
	\caption{Comparison of average episode return as a function of the environment steps. The solid lines demonstrate the average performance over eight random seeds while the shaded areas depict the standard deviation from the average. Finally, the dashed lines stand for the best performance of the demonstrator.}
	\label{fig:eps return}
\end{figure*}

\subsection{Experiment Setup}
Assume we selected \textit{Assault} as the testing task, then the remaining five games were set as the training tasks. For training task, we subsampled $1000$ trajectory pairs by random selection, and $80\%$ of the pairs were used as the support set. For testing task, we subsampled $500$ trajectory pairs by random selection, and $80\%$ of the pairs were also used as the support set.

The first step is to train the parameterized reward function on the derived demonstrations. As shown in Table \ref{tb:cnn na}, $\hat{R}_{\bm\theta}$ has three convolutional layers and two fully-connected layers, and each convolutional layer is followed by a batch normalization layer. Furthermore, ReLU is used as the activation function. Note that "$8\times8$ Conv. $32$" represents a convolutional layer that has $32$ filters of size $8\times8$. To reduce the computational complexity, we propose to stack four consecutive frames as an input before resizing the input into patches of size $(84, 84)$.

In the first iteration step, we sampled a training task $\mathcal{T}$ and built an identical reward network for it. After that, $\mathcal{T}$ conducted training on its support set, in which an Adam optimizer with a learning rate of $\alpha=0.0005$ was used to perform gradient descent-based updates. Next, we calculated the test loss with its query set followed by the gradient computation with respect to the updated parameters, and updated $\hat{R}_{\bm\theta}$ using an SGD optimizer with a learning rate of $\beta=0.0001$. We repeated the procedures above for $100$ times, and saved the model weights for the subsequent fine-tuning procedure. Equipped with the parameters learned from the previous stage, we continued training the reward function using the support set of the testing task, and the number of epoch was set to $100$. After that, the reward function was saved to perform policy optimization.

For the policy update, we used a PPO method with a learning rate of $0.0025$, a value function coefficient of $0.5$, an entropy coefficient of $0.01$, and a generalized advantage estimation (GAE) parameter of $0.95$. In particular, a gradient clipping operation with threshold $[-5,5]$ was performed to stabilize the learning procedure. To make a fair comparison, we used an identical policy network and a value network for all methods. The detailed architectures are illustrated in Table \ref{tb:cnn na}. For benchmarking schemes, we trained them following the default configurations reported in their literature \cite{yu2020intrinsic, brown2019extrapolating, xiao2019wasserstein}.

\begin{table*}[ht]
	\centering
	\caption{Average return comparison in Atari games.}
	\label{tb:atari results}
	\begin{tabular}{l|ll|llll}
		\hline
		\multicolumn{1}{c|}{\multirow{2}{*}{Game}} & \multicolumn{2}{c|}{Demonstrations}   & \multicolumn{4}{c}{Algorithms}                                                                                                            \\ \cline{2-7}
		\multicolumn{1}{c|}{}                      & \multicolumn{1}{l|}{Best}   & Average & \multicolumn{1}{l|}{MLRE}              & \multicolumn{1}{l|}{DREX}              & \multicolumn{1}{l|}{GIRIL}            & WAIL            \\ \hline
		Assault                                    & \multicolumn{1}{l|}{3.94k}  & 3.41k   & \multicolumn{1}{l|}{$\bf 4.56k\pm1.88k$}   & \multicolumn{1}{l|}{3.89k$\pm$1.49k}   & \multicolumn{1}{l|}{3.72k$\pm$1.61k}  & 0.8k$\pm$0.17k  \\
		Battle Zone                                & \multicolumn{1}{l|}{19.76k} & 17.72k  & \multicolumn{1}{l|}{$\bf 21.38k\pm5.59k$}  & \multicolumn{1}{l|}{$\bf 21.00k\pm8.11k$}   & \multicolumn{1}{l|}{19.62k$\pm$7.3k}  & 3.75k$\pm$1.57k \\
		Kung Fu Master                             & \multicolumn{1}{l|}{23.59k} & 11.63k  & \multicolumn{1}{l|}{$\bf 28.02k\pm11.92k$} & \multicolumn{1}{l|}{$\bf 27.04k\pm11.52k$} & \multicolumn{1}{l|}{16.95k$\pm$4.15k} & 1.96k$\pm$0.47k \\
		Phoenix                                    & \multicolumn{1}{l|}{7.87k}  & 6.34k   & \multicolumn{1}{l|}{$\bf 8.89k\pm3.25k$}   & \multicolumn{1}{l|}{$\bf 8.81k\pm3.13k$}   & \multicolumn{1}{l|}{$\bf 8.46k\pm1.83k$}  & 1.97k$\pm$0.55k \\
		Riverraid                                  & \multicolumn{1}{l|}{8.85k}  & 7.47k   & \multicolumn{1}{l|}{$\bf 9.74k\pm0.9k$}    & \multicolumn{1}{l|}{8.72k$\pm$1.3k}    & \multicolumn{1}{l|}{$\bf 9.57k\pm1.13k$}  & 1.63k$\pm$0.29k \\
		Space Invaders                             & \multicolumn{1}{l|}{0.82k}  & 0.65k   & \multicolumn{1}{l|}{$\bf 1.06k\pm0.29k$}   & \multicolumn{1}{l|}{$\bf 0.85k\pm0.35k$}  & \multicolumn{1}{l|}{$\bf 0.86k\pm0.41k$}  & 0.26k$\pm$0.09k \\ \hline
	\end{tabular}
\end{table*}

 \subsection{Results}
\subsubsection{Reward Extrapolation}
We first investigated the capability of the learned reward function via MLRE and DREX. To that end, we compared the ground-truth return and the inferred return of MLRE on multiple collected trajectories, and the results are shown in Fig.~\ref{fig:re}. For \textit{Assault}, MLRE performed considerable prediction in the whole performance range, while DREX produced large variance for ground-truth high returns. For \textit{Phoenix}, the predicted returns of DREX were always higher than that the ground-truth returns when the demonstrations had shorter lengths. Finally, both MLRE and DREX made reasonably good predictions for \textit{Space Invaders}. But MLRE achieved higher prediction accuracy.

\subsubsection{Policy Performance}
For performance comparison, the average one-life return is utilized as the key performance indicator (KPI). Table \ref{tb:atari results} illustrates the performance comparison over eight random seeds, in which the beyond-demonstrator performance is shown in bold. MLRE outperformed the best demonstration in all the six games, achieving an average performance gain of $15.8\%$. DREX and GIRIL outperformed the best demonstration in four and three games, producing an average performance gain of $10.9\%$ and $6.83\%$, respectively. In comparison, WAIL performed worse than the average performance of the demonstrations in all games. Despite abundant training data, WAIL performed poorly in complex environments with high-dimensional observations. Finally, we provide detailed learning curves of all the games Fig.~\ref{fig:eps return}. It is obvious that BDIL algorithms realized stable and rapid performance growth, while the IL algorithm is futile with limited training data.

\section{CONCLUSION}
In this paper, we have investigated the problem of beyond-demonstrator imitation learning, and proposed a meta learning-based reward extrapolation framework entitled MLRE. By exploiting the meta learning mechanism, MLRE can learn high-quality reward functions even for limited demonstrations, which makes MLRE attractive for real-world applications. Extensive simulation using multiple Atari games was performed to confirm that MLRE outperforms the existing BDIL algorithms with higher efficiency and robustness.


\end{document}